\pdfoutput=1
\documentclass[11pt]{article}

\usepackage[preprint]{acl}

\usepackage{times}
\usepackage{latexsym}
\usepackage{graphicx}
\usepackage{amsmath}
\usepackage{multirow}
\usepackage{multicol}
\usepackage{tabularx}
\usepackage{makecell} 
\usepackage{bm}
\usepackage{colortbl}  
\usepackage{stfloats}
\usepackage{wrapfig}
\usepackage{afterpage}
\usepackage{algorithm}
\usepackage{algorithmic}
\usepackage{arydshln}
\usepackage{subcaption}

\usepackage[T1]{fontenc}
\usepackage[utf8]{inputenc}

\usepackage{microtype}

\usepackage{inconsolata}

\usepackage{amsfonts}
\usepackage{comment}
\usepackage{amsmath}
\usepackage{float}
\usepackage{multicol}

\title{Progressive Mastery: Customized Curriculum Learning with Guided Prompting for Mathematical Reasoning}

\author{Muling Wu\textsuperscript{1}\thanks{These authors contributed equally}, Qi Qian\textsuperscript{1}\footnotemark[1], Wenhao Liu\textsuperscript{1}, Xiaohua Wang\textsuperscript{1}, Zisu Huang\textsuperscript{1} \\
{\bf Di Liang\textsuperscript{2}, LI Miao\textsuperscript{2}, Shihan Dou\textsuperscript{1}, Changze Lv\textsuperscript{1}, Zhenghua Wang\textsuperscript{1}} \\
 {\bf Zhibo Xu\textsuperscript{1}, Lina Chen\textsuperscript{1}, Tianlong Li\textsuperscript{1}, Xiaoqing Zheng\textsuperscript {1}\thanks{Corresponding author.}, Xuanjing Huang\textsuperscript{1} }\\
  \texttt{\{mlwu22,qqian23\}@m.fudan.edu.cn} \\
   \textsuperscript{1}Fudan University \textsuperscript{2}ByteDance Inc.\\
}

\begin{document}
\maketitle

\begin{abstract}
Large Language Models (LLMs) have achieved remarkable performance across various reasoning tasks, yet post-training is constrained by inefficient sample utilization and inflexible difficulty samples processing.
To address these limitations, we propose Customized Curriculum Learning (CCL), a novel framework with two key innovations. First, we introduce model-adaptive difficulty definition that customizes curriculum datasets based on each model's individual capabilities rather than using predefined difficulty metrics. Second, we develop "Guided Prompting," which dynamically reduces sample difficulty through strategic hints, enabling effective utilization of challenging samples that would otherwise degrade performance.
Comprehensive experiments on supervised fine-tuning and reinforcement learning demonstrate that CCL significantly outperforms uniform training approaches across five mathematical reasoning benchmarks, confirming its effectiveness across both paradigms in enhancing sample utilization and model performance.

\end{abstract}
\section{Introduction}
Large Language Models (LLMs) have achieved breakthrough progress in the field of natural language processing (NLP) in recent years. 
With additional post-training optimization, these models have demonstrated exceptional performance on complex tasks such as code generation and mathematical reasoning\cite{Guo2024DeepSeekCoderWT,Shao2024DeepSeekMathPT,Yang2025Qwen3TR,gemini25,openaio1}. 
However, the current post-training process for LLMs still faces significant challenges. 
One notable limitation of conventional training is the uniform treatment of all examples, ignoring their varying difficulty levels or value. 
Consequently, challenging or high-quality samples are not strategically introduced at optimal points in the training process, thereby impeding effective knowledge acquisition and integration.

To address this limitation, \citet{Bengio2009CurriculumL} introduced the concept of curriculum learning to model training, inspired by human education's progression from simple to complex concepts.
Several recent studies have also begun to explore the application of curriculum learning strategies based on heuristic rules to the LLM post-training pipeline.
For instance, in logical reasoning task, \citet{Xie2025LogicRLUL} measured example difficulty by input length to implement progressive training from simpler to more complex instances. 
\citet{Wen2025LightR1CS} classified difficult examples as those incorrectly predicted by DeepSeek-R1\cite{DeepSeekAI2025DeepSeekR1IR},and prioritized these challenging cases during later training stages.  
Although these rule-based curriculum learning methods improved performance to some extent, they nonetheless present certain limitations.

First, these predefined difficulty metrics lack precision in measuring actual difficulty levels. 
As illustrated in Figure \ref{fig:difficult_level}, our experiments on the MATH dataset demonstrate that model performance does not consistently decline with increasing predefined difficulty levels. 
Counterintuitively, models achieve higher accuracy on purportedly more difficulty Level 5 problems than on Level 4 problems.
Second, defining difficulty using a uniform standard proves inadequate, as metrics that effectively gauge difficulty for one model often fail to appropriately characterize challenge levels for another model.
As illustrated in Figure \ref{fig:model_comparison}, samples that present significant challenges for Qwen2.5-MATH-7B are often solved with ease by DeepSeek-Math-7B-Instruct, and vice versa. 
To address these limitations, we propose a tailored curriculum learning approach that calibrates sample difficulty according to each model's individual capabilities, thereby customizing more appropriate training sequences for different models.

Another significant challenge in model training stems from the presence of extremely challenging examples in the training data. 
Previous research \cite{Yu2025DAPOAO,Wen2025SARISA} has demonstrated that forcing models to train on examples substantially beyond their current capabilities can lead to performance degradation. 
Consequently, a conventional approach has been to simply exclude such overly difficult samples from the training process to prevent negative impacts on model learning.
However, this wholesale elimination of challenging data is inherently inefficient, as these difficult examples often contain valuable information that could potentially enhance model performance if leveraged appropriately.
To overcome this limitation, we introduce "Guided Prompting," a technique that augments input examples with targeted hints to dynamically modulate their difficulty during the training process.
This method effectively prevents performance deterioration while enabling the model to extract meaningful patterns from examples that would otherwise be discarded, thereby significantly improving overall data utilization efficiency.

The contributions of this study are summarized as follows:
\vspace{-2mm}
\begin{itemize}
\setlength{\itemsep}{0pt}
\setlength{\parsep}{0pt}
\setlength{\parskip}{0pt}
\item We tailored the course dataset based on the model's performance and proposed a novel post-training approach called Customized Curriculum Learning (CCL).
\item We implement "Guided Prompting" for samples that significantly exceed current model capabilities, effectively controlling sample difficulty and substantially improving data utilization efficiency.
\item We conduct comprehensive experiments across two mainstream post-training paradigms, namely supervised fine tuning and reinforcement learning, demonstrating significant performance improvements on our specific model.
\end{itemize}

\section{Related Work}
\textbf{Curriculum Learning.} 
\citet{Bengio2009CurriculumL} introduces the concept of curriculum learning, demonstrating that models learn more effectively when training examples are presented in a progressively harder order.
Recent approaches in large language models build on this concept.
\citet{Xie2025LogicRLUL} designs curricula by adjusting task difficulty based on logical complexity, enhancing the model’s reasoning abilities.
\citet{Wen2025LightR1CS} treats queries that DeepSeek-R1\cite{DeepSeekAI2025DeepSeekR1IR} struggles with as hard samples, deferring them to later training stages for focused learning.
\citet{Team2025KimiKS} refine training by filtering out noisy samples early, concentrating on high-quality examples for later stages.
\citet{Huang2025RAGRLAR} apply curriculum learning to retrieval-augmented generation (RAG), ordering tasks based on the number of distractors in retrieved passages. 
\citet{Shi2025EfficientRF} dynamically selects training samples whose predefined difficulty scores are closest to the model’s current target difficulty level, which is adjusted during training based on reward feedback.
Unlike these approaches that rely on heuristic rules to define a fixed difficulty hierarchy shared across all models, we propose a customized curriculum learning framework that tailors the training sequence to each model’s reasoning ability, enabling a more adaptive and effective learning process.

\noindent\textbf{Model-Adaptive Difficulty Awaring.}
In the model training process, different types of samples should be treated with varying degrees of emphasis, with difficulty-aware methods serving as a key approach to distinguishing data types.
\citet{sky_t1_2025,Xie2024GradeLA,10673924} employ LLM-generated scoring to assess sample difficulty, while 
\citet{Min2024ImitateEA,Yuan2025AgentRTL,Wen2025LightR1CS}  heuristically treat long-form QA tasks as inherently challenging. 
\citet{Tong2024DARTMathDR,Xue2025DASTDS} take a more empirical approach by conducting multiple sampling iterations for each query, defining difficulty through incorrect response ratios and allocating more trials to challenging queries during synthesis. Similarly, \citet{Ma2024PlugandPlayTF} implements multi-round query sampling but weights samples inversely proportional to accuracy, thereby giving higher weights to samples with lower accuracy scores.

\noindent\textbf{Post Training.}
Supervised fine-tuning and reinforcement learning represent the two most prevalent methods in post-training.
Fine-tuning pretrained models on high-quality datasets with step-by-step solutions markedly enhances problem-solving accuracy\cite{Yue2023MAmmoTHBM,Yuan2023ScalingRO,Hwang2024SelfExploreEM}.
Beyond supervised learning, reinforcement-based fine-tuning has been explored to further align LLMs with solution correctness and preferred reasoning styles. 
\citet{Luo2023WizardMathEM,Luong2024ReFTRW,Yue2025VAPOEA} optimize policy networks using Proximal Policy Optimization (PPO). 
In contrast, \citet{Shao2024DeepSeekMathPT,DeepSeekAI2025DeepSeekR1IR,Yu2025DAPOAO}  replace the critic model in PPO and optimize policy networks via Group Relative Policy Optimization (GRPO).

\begin{figure*}[!t]
  \centering
  \includegraphics[width=\linewidth]{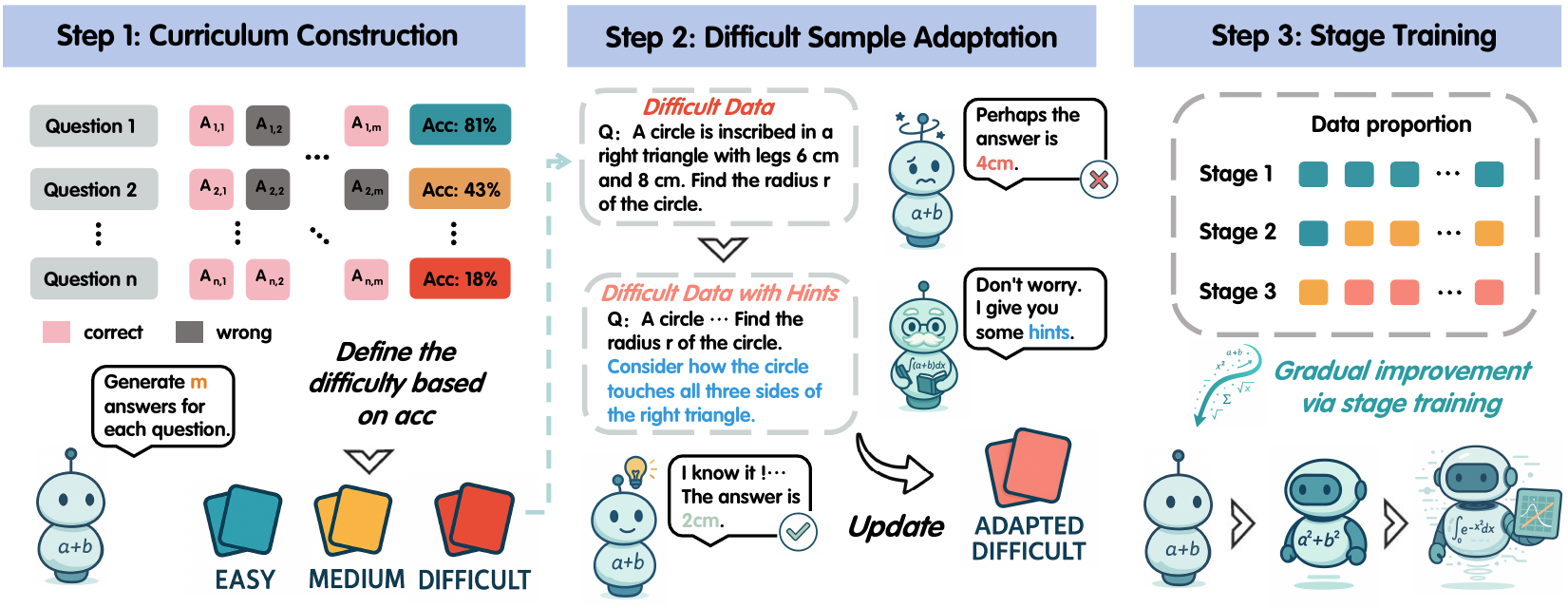}
  \caption{\textbf{Overall pipeline of our method.}
\textbf{Step 1:} For each question in the training set, the model generates multiple responses and calculates the accuracy on that sample. Based on these accuracy scores, samples are ranked and organized into curriculum datasets.
\textbf{Step 2:} Transforming difficult samples to reduce the answering difficulty for the model, bringing samples within the model's solvable range.
\textbf{Step 3:} The model undergoes staged training sequentially on easy, medium, and difficult curriculum datasets, with its performance continuously enhanced.}
\label{fig:workflow}
\end{figure*}

\section{Method}
Traditional training approaches treat all samples equally, failing to adequately leverage high-quality samples, which leads to suboptimal performance. 
To address this limitation, we propose the Customized Curriculum Learning (CCL) training framework, which enables models to learn progressively from easy to difficult samples. 
By structuring the training process to prioritize foundational concepts before advancing to more complex samples, CCL enables models to build extremely solid foundations during the early stages of training, upon which more sophisticated understanding can be constructed.

The CCL training framework consists of three key steps. First is curriculum construction. 
Since curriculum learning proceeds in stages across samples of varying difficulty, the original dataset must be partitioned accordingly. We propose an aptitude-based training approach that customizes curriculum for each model according to its inherent capabilities, applying the principle of individualized instruction to optimize learning progression. Second is difficult sample adaptation. Due to inherent model limitations, some samples remain consistently challenging regardless of the model's attempts. Training on such data can actually degrade model performance. We identify these problematic samples and implement a "Guided Prompting" method to reduce the answering difficulty, thereby improving sample utilization efficiency. The final step is multi-stage training. Utilizing the constructed and modified curriculum datasets from previous steps, we implement staged supervised fine-tuning and reinforcement learning, enabling the model to gradually adapt to samples of increasing difficulty, enhancing training stability and overall performance, the full algorithm is detailed in Algorithm \ref{alg:ccl}.

\subsection{Curriculum Construction}
\begin{figure}[t]
  \centering
  \includegraphics[width=\linewidth]{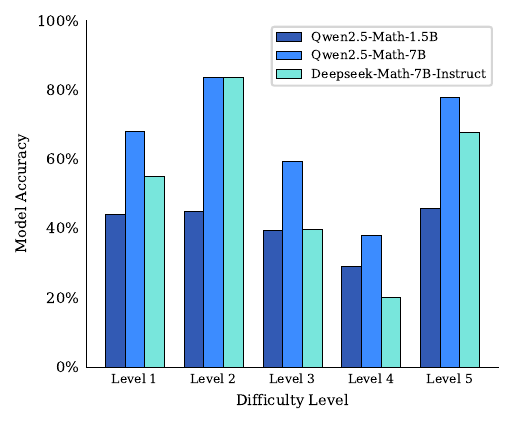}
\caption{Performance of multiple models on MATH dataset subsets with predefined difficulty levels. As predefined difficulty increases from Level 1 to Level 5, model accuracy does not consistently decline but instead exhibits significant fluctuations, demonstrating that predefined difficulty standards may not correctly adapt to all models. }  
  \label{fig:difficult_level}
  \vspace{-4mm}
\end{figure}

\begin{figure}[H]
  \centering
  \includegraphics[width=\linewidth]{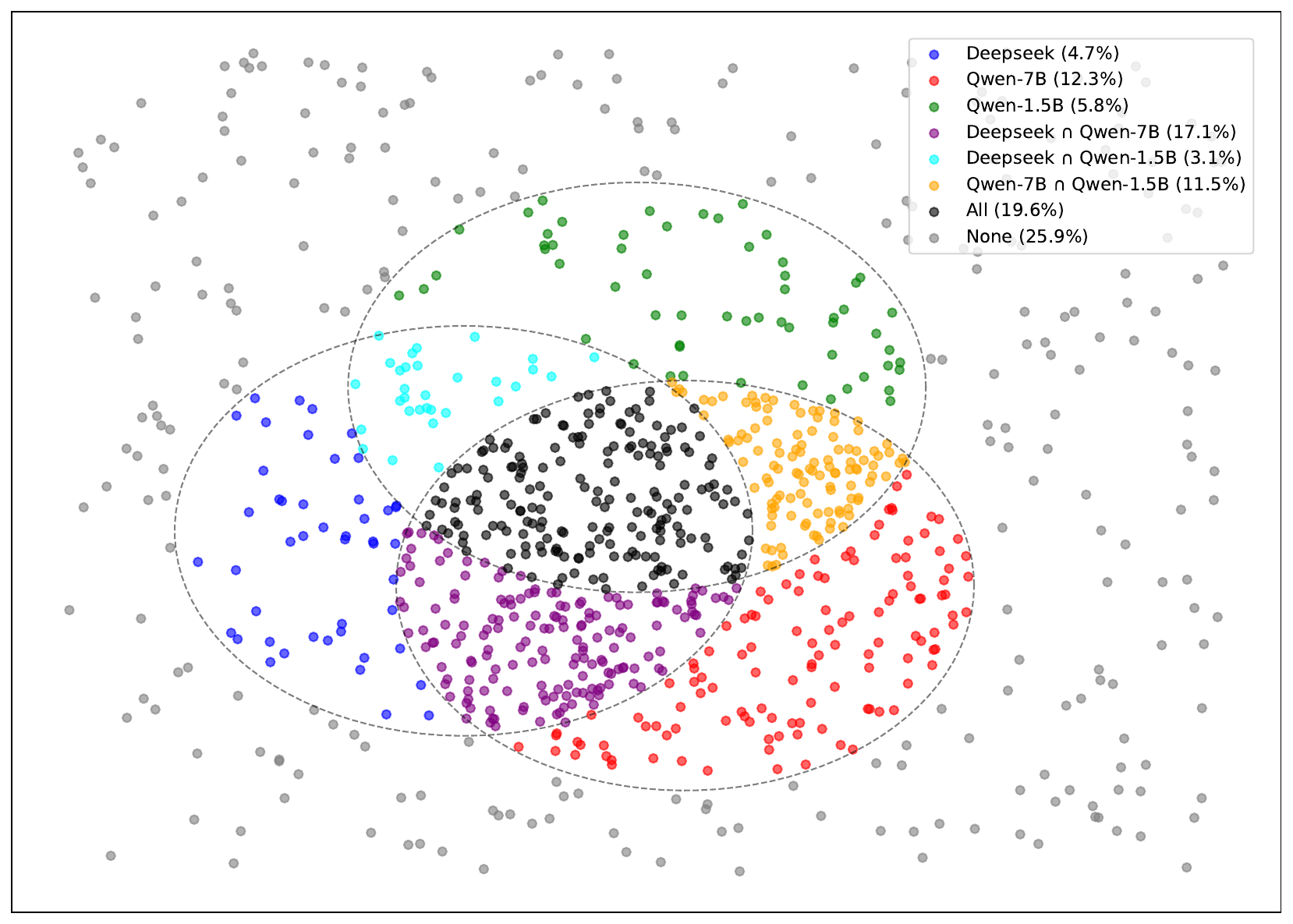} 
\caption{
Visualization of solution correctness patterns for three mathematical reasoning models: 
Deepseek-Math-7B-Instruct (\textcolor{blue}{blue}), 
Qwen2.5-Math-7B (\textcolor{red}{red}), 
and Qwen2.5-Math-1.5B (\textcolor{green!50!black}{green}). 
Each colored region represents a specific answering scenario; for example, the black region indicates questions all three models answered correctly.
Approximately $55$\% of questions that are easy for one model prove difficult for another, demonstrating that using a unified standard to define sample difficulty across all models is unreasonable.
}

  \label{fig:model_comparison}
  \vspace{-4mm}
\end{figure}

To implement training in an easy-to-difficult sequence, we first need to establish metrics for measuring sample difficulty, which will then be used to sort and segment the dataset. \citet{Ding2024MitigatingTN} directly used the manually annotated difficulty levels in the MATH dataset as the standard for distinguishing sample complexity. However, this heuristic definition has certain limitations. As shown in Figure \ref{fig:difficult_level}, we selected Qwen2.5-Math-1.5B\cite{Yang2024Qwen25MathTR}, Qwen2.5-Math-7B, and Deepseek-Math-7B\cite{Shao2024DeepSeekMathPT} as baseline models and tested them on the MATH dataset. 
The results indicate that model performance does not consistently decrease as the predefined difficulty level increases (for instance, models achieve higher accuracy on level 4 problems than on level 5 problems), suggesting that this heuristic definition lacks precision. 
Furthermore, Figure \ref{fig:model_comparison} demonstrates that defining difficulty using a uniform standard proves inadequate. Samples that are extremely simple for Qwen2.5-Math-1.5B may still challenge Qwen2.5-Math-7B, and vice versa, indicating that we cannot establish a single unified difficulty measurement standard that is suitable for all models.

To address these two issues, we propose a "teaching according to aptitude" curriculum construction approach, which customizes curriculum datasets based on each model's own performance. Specifically, we input all training samples into the model for inference and collect the model's responses through sampling. For each problem, the model provides N responses, and the model's accuracy rate for that sample is defined as following:
\begin{equation}
    ACC_{i} = \frac{\sum_{j=1}^{n}{\mathbf{1}\{A_{ij}=A_i^*\}}}{n}
\end{equation}
where $ACC_i$ denotes the accuracy of $i$-th samples $A_{ij}$ denotes the response generated by the model for the $i$-th sample in the $j$-th sampling iteration, and $A_i^*$ denotes the golden answer of $i$-th samples.
We define the difficulty level of a sample as the inverse of this accuracy rate and sort the samples according to their difficulty levels, as shown in Figure\ref{fig:workflow} Step 1. Samples that the model can correctly answer multiple times are classified as simple samples and can be well mastered in the early stages of training. Samples that the model repeatedly fails to solve are classified as moderate or difficult samples, which are reserved for later stages of training after the model has acquired the relevant foundational knowledge in the domain, allowing it to more thoroughly learn these more complex samples.

\begin{algorithm*}[t]
\caption{Customized Curriculum Learning for Enhancing Mathematical Reasoning}
\begin{algorithmic}[1]
\REQUIRE training dataset $D = \{(Q_i, A_i)\}$ with questions $Q_i$ and reference solution $A_i$
\REQUIRE pretrained model $\pi_0$, accuracy threshold $\tau$, hint ratio $\alpha$
% \ENSURE Updated model $\pi_\theta$ with enhanced capability on mathematical reasoning tasks

\STATE \textbf{Curriculum Construction:}
\FORALL {question $Q_i \in D$}
    \STATE Generate $n$ response $\{A_{i1}, \ldots, A_{in}\}$ using model $\pi_\theta$
    \STATE Calculate $ACC_i = \frac{\sum_{j=1}^{n}{\mathbf{1}\{A_{ij}=A_i^*\}}}{n}$, where $A_i^*$ is the golden answer of $Q_i$
\ENDFOR
\STATE Sort all samples in descending order based on their $ACC_i$ values, then partition the dataset $D$ into $p$ disjoint subsets $\{D_1, D_2,...,D_p\}$, where the $p$-th subset $D_p$ contains samples with the lowest $ACC_i$ values, representing the most challenging instances.

\STATE \textbf{Difficult Sample Adaptive Processing:}
\FORALL {difficult samples $(Q_i, A_i) \in D_p$}
    \STATE Decompose the reference solution $S_i$ into a sequence of problem-solving steps $\{s_{i1}, \ldots, s_{ik}\}$
    \STATE Gradually provide hints $P_i = \{s_{i1}, \ldots, s_{il}\}$ until either the ratio $\frac{|il|}{|ik|}$ reaches $\alpha$, or the model's performance improves to the $\tau$
   \IF{model's performance improves to $\tau$}
    \STATE Update question and answer: $Q_i \rightarrow [Q_i; P_i]$, $A_i \rightarrow \{s_{i(l+1)}, \ldots, s_{ik}\}$
\ELSE
    \STATE Discard this overly difficult sample $(Q_i, A_i)$ that the model still fails to handle effectively even with hints
\ENDIF
\ENDFOR

\STATE \textbf{Multi-Stage Training Process:}
\FOR{each stage $s \in \{1,\ldots,p\}$}
    \STATE Fine-tune the previous stage’s model $\pi_{s-1}$ on current stage dataset $D_s$:
    \[
    \pi_{s} = \arg\min_{(Q,A) \in D_{s}} \mathcal{L}(\pi_{s-1})
    \]
\ENDFOR

\STATE \textbf{Output:} Enhanced model $\pi_m$ with improved problem-solving skills
\end{algorithmic}
\label{alg:ccl}
\end{algorithm*}

\subsection{Difficult Sample Adaptation}
Even after constructing curriculum data from easy to difficult as outlined in step 1, there remain samples that the model cannot solve correctly regardless of how many attempts it makes, due to limitations in the model's inherent capabilities. 
\citet{Yu2025DAPOAO,Wen2025SARISA} have demonstrated that training on samples far beyond the model's current capabilities actually degrades performance, leading to the common practice of discarding such samples. However, we consider direct disposal of this data to be wasteful and propose a "guided prompting" method to transform and reclaim these difficult samples.

Similar to how teachers approach difficult concepts in education, we apply a pedagogical framework to model training. When students encounter obstacles while learning new material, effective teachers don't skip challenging concepts simply because they exceed the student's current understanding. Instead, they employ progressive, step-by-step guidance to facilitate knowledge absorption. 
By incorporating this educational philosophy into model training, we provide the model with hints that guide its solution generation process\cite{Xi2024TrainingLL,Dou2025ImprovingRE}. 
This significantly reduces the difficulty of problem-solving and helps the model better assimilate the current knowledge, as illustrated in Figure \ref{fig:workflow} Step 2.

Specifically, for a problem $Q_i$ with corresponding reference answer $S_i$, we first decompose $S_i$ into step-by-step solution components, such that $S_i = \{s_{i1}, s_{i2},...,s_{ik}\}$. 
We then extract a small prefix $P_i = \{s_{i1}, s_{i2},...,s_{ip}\}$ from $S_i$ to serve as a hint, where $p$ is smaller then $k$.
This prefix $P_i$ is concatenated with $Q_i$ as input to guide the model toward generating the correct answer, that is
\begin{equation}
    y_i \sim \pi_\theta(Y|[Q_i;P_i])
\end{equation}
where $y_i$ denotes response generated by model $\theta$.
Through this approach, samples that previously exceeded the model's capabilities are transformed into manageable examples. 
The originally complex answer generation task becomes a simpler answer completion task, thereby enhancing the model's effective utilization of training samples.

\subsection{Multi Stage Training}
After constructing the curriculum dataset and modifying the difficult samples, we can utilize this partitioned data $D = \{D_1, D_2,...,D_p\}$ for multi stage SFT and multi stage RL, as shown at Figure \ref{fig:workflow} Step 3.
For model $\pi_{\theta}$, dataset $D_j$ has a higher difficulty level than dataset $D_i$, where $j$>$i$.
\subsubsection{Multi Stage SFT}
Let the pre-trained model be denoted as $\pi_0$. Based on the number of partitions in the curriculum dataset, multi-stage SFT necessitates $m$ rounds of maximum likelihood optimization. 
The loss function for the $i$-th round of optimization can be formally expressed as follows:
\begin{align}
    \mathcal{L}_{SFT} = -\mathbb{E}_{(x,y) \sim D_i}[log(\pi_{i-1}(y|x))]
\end{align}
\subsubsection{Multi Stage RL}
Based on the constructed curriculum dataset, we introduce multi-stage reinforcement learning to enhance the model's generalizability. 
\citet{DeepSeekAI2025DeepSeekR1IR} has demonstrated that reinforcement learning can be performed directly without supervised fine-tuning, known as DeepSeek-R1-Zero, which can also achieve excellent performance on reasoning tasks. We follow this setting to conduct our experiments.

The reward is the source of training signals in reinforcement learning and determines the optimization direction of the entire reinforcement learning process. 
To effectively train reasoning models through reinforcement learning, following the setup in \cite{DeepSeekAI2025DeepSeekR1IR}, we adopt a rule-based reward function that includes two types of rewards. 
The first is the format reward, which measures whether the model outputs according to our required format. If the response contains special tokens such as "<think>,</think>,<answer>,</answer>", it is considered that the format is correct, formally represented as follows: 
\begin{align}
r_{format} &=
  \begin{cases}
    1.0 & \text{format is correct} \\
    0.0 & \text{others}
  \end{cases} 
\end{align}
The second type is the accuracy reward, which measures whether the model's final prediction is correct. 
We extract the model's prediction from the generated response according to the rules and compare it with the golden answer, formally represented as follows:
\begin{align}
r_{accuracy} &=
  \begin{cases}
    1.0 & \text{prediction is correct} \\
    0.0 & \text{others}
  \end{cases}
\end{align}
The final reward $r$ equals the sum of both, that is $r = r_{format} + r_{accuracy}$.

We employ Group Relative Policy Optimization (GRPO) as our policy learning algorithm\cite{Shao2024DeepSeekMathPT}. 
The GRPO algorithm generates multiple candidate responses  $O$ for each question $Q$, where $O=\{o_1,o_2,...,o_G\}$. These different responses to the same question form a group, and the reward of each response in this group is used to calculate the advantage $A_i$ of that sample, as follows:
\begin{equation}
    A_i = \frac{r_i-maen(\{r_1,r_2,...,r_G\})}{std(\{r_1,r_2,...,r_G\})}
\end{equation}
GRPO adopts a clipped objective, together with a directly imposed KL penalty term:
\begin{align}
    \mathcal{L}_{GRPO} = \mathbb{E}_{(x) \sim D}[\frac{1}{G}\sum_{i=1}^{G}\frac{1}{|o_i|}\sum_{t=1}^{|o_i|}(min(r_{i,t}(\theta)A_i, \nonumber\\
    clip(r_{i,t}(\theta), 1-\epsilon, 1+\epsilon)A_i)-\beta D_{KL}(\pi_{\theta}||\pi_{ref}))], \nonumber\\
\end{align}
where
\begin{align}
    r_{i,t}(\theta) = \frac{\pi_{\theta}(o_{i,t}|q,o_{i,<t})}{\pi_{\theta_{old}}(o_{i,t}|q,o_{i,<t})}
\end{align}

% 调整行间距
\renewcommand{\arraystretch}{1.75}
\setlength{\tabcolsep}{6pt}

\begin{table*}[htbp]
\centering
\small
\resizebox{\textwidth}{!}{
\begin{tabular}{
    l  % Model
    c  % Method
    l  % Stage
    c  % MATH 500
    c  % Minerva Math
    c  % Olympiad Bench
    c  % AIME24
    c  % AMC23
    c  % Average
}
\hline
\makecell[l]{\rule{0pt}{1.2em}\textbf{Model}} 
& \makecell[l]{\rule{0pt}{1.2em}\textbf{Method}} 
& \makecell[c]{\rule{0pt}{1.2em}\textbf{Learning} \\ \textbf{Strategy}} 
& \makecell{\rule{0pt}{1.2em}\textbf{MATH  } \\ \textbf{500}} 
& \makecell{\rule{0pt}{1.2em}\textbf{Minerva} \\ \textbf{Math}} 
& \makecell{\rule{0pt}{1.2em}\textbf{Olympiad} \\ \textbf{Bench}} 
& \makecell{\rule{0pt}{1.2em}\textbf{AIME24}} 
& \makecell{\rule{0pt}{1.2em}\textbf{AMC23}} 
& \makecell{\rule{0pt}{1.2em}\textbf{Average}} \\
\hline

\multirow{4}{*}{\textbf{Qwen2.5-Math-1.5B}} 
 & \multirow{2}{*}{SFT} & \textit{Uniform} & $\bm{48.60}$ & $18.00$ & $12.60$ & $0.00$ & $27.50$ & $21.34$ \\   
  & & \textit{CCL} & $48.00$ & $\bm{23.50}$ & $\bm{12.90}$ & $0.00$ & $27.50$ & $\bm{22.38}$ \\ \cline{2-9} 
 & \multirow{2}{*}{GRPO} & \textit{Uniform} & $51.80$ & $18.40$ & $21.00$ & $10.00$ & $22.50$ & $24.74$ \\  
  & & \textit{CCL} & $\bm{72.60}$ & $\bm{31.60}$ & $\bm{32.70}$ & $\bm{13.30}$ & $\bm{42.50}$ & $\bm{38.54}$ \\ 
\hline

\multirow{4}{*}{\textbf{Qwen2.5-Math-7B}}  
 & \multirow{2}{*}{SFT} & \textit{Uniform} & $\bm{68.80}$ & $16.50$ & $18.40$ & $0.00$ & $22.50$ & $25.24$ \\  
 & & \textit{CCL} & $63.00$ & $\bm{21.00}$ & $\bm{18.70}$ & $\bm{3.30}$ & $\bm{45.00}$ & $\bm{30.20}$ \\ \cline{2-9}

 & \multirow{2}{*}{GRPO} & \textit{Uniform} & $74.20$ & $33.50$ & $33.90$ & $10.00$ & $\bm{62.50}$ & $42.82$ \\  
  & & \textit{CCL} & $\bm{76.60}$ & $\bm{38.20}$ & $\bm{38.20}$ & $\bm{13.30}$ & $60.00$ & $\bm{45.26}$ \\
\hline

\end{tabular}
}
% \caption{Performance Results for Different Models and Methods on Benchmark Math Datasets}
\caption{Evaluation results of different learning strategies on Math Datasets}
\label{tab:main-results}%
\end{table*}

\vspace{1em} % 在表格后增加1em的空白

\section{Experiments}
\subsection{Datasets}
\noindent\textbf{Train.}
Following the experimental setting of \cite{zeng2025simplerlzooinvestigatingtamingzero}, we selected the MATH dataset\cite{Hendrycks2021MeasuringMP} and extracted samples from level $3$ to level $5$ as training data, comprising a total of $9,255$ instances. To adapt our proposed CCL framework for creating a customized curriculum dataset for the model, we first needed to differentiate these samples based on their difficulty levels according to the model's performance on them.
After completing inference on all samples, we ranked them according to the model's accuracy rate. 
Samples with higher accuracy rates were categorized as simple data for use in the early stages of model training, while samples with lower accuracy rates were designated as difficult data for later training stages.
See Appendix \ref{appendix:data} for more detailed descriptions.
Additionally, the data was processed into a conversational format. 
The prompts used in the SFT and GRPO processes can be found at Appendix \ref{appendix:prompt}.

\noindent\textbf{Test.} We use five benchmark datasets to assess the model’s performance across different
levels of difficulty and mathematical reasoning. 
MATH 500\cite{Lightman2023LetsVS}, is a subset of the MATH dataset, containing $500$ representative problems designed to test a model’s general mathematical capability.
OlympiadBench \cite{He2024OlympiadBenchAC} includes a collection of problems from Olympiad-level mathematics and physics competitions. 
Minerva Math \cite{Lewkowycz2022SolvingQR} is a curated set of undergraduate-level math problems that assess complex mathematical
reasoning and symbolic manipulation. 
AMC 23 and AIME 24 include problems from the 2023 American Mathematics Competitions and the 2024 American Invitational Mathematics Examination, respectively.
Additionally, the data was processed into a conversational format. 
 
\subsection{Models}
To effectively validate the efficacy of our CCL method across foundation models of varying capabilities, we selected two different-sized models for our experiments: Qwen2.5-MATH-1.5B\cite{Yang2024Qwen25MathTR} and Qwen2.5-MATH-7B.

\subsection{Training Setup}
We conducted our experiments using $8$ NVIDIA A100 GPUs for the SFT experiments within the Llama-Factory framework\cite{zheng2024llamafactory} and for the GRPO experiments within Hugging Face's Open R1 framework\cite{openr1}.
See Appendix \ref{appendix:hp} for more detailed descriptions.

\subsection{Evaluation Setup}
We evaluated our models using the evaluation script from \cite{zeng2025simplerlzooinvestigatingtamingzero}.
See Appendix \ref{appendix:eval} for more detailed descriptions.

\subsection{Main Results}
We implemented various strategies and training methods across multiple models of different scales and conducted extensive experiments on test sets of varying difficulty levels.
As shown in Table \ref{tab:main-results}, compared to uniform training that treats all data equally, our CCL strategy demonstrates significant advantages by customizing curriculum datasets and adopting an easy-to-hard training approach, yielding substantial performance improvements across multiple experimental settings.

Our CCL learning strategy demonstrates both method compatibility and model scalability. When applied to different-sized models under SFT settings, CCL improved performance by $1.04\%$ and $4.96\%$ for Qwen2.5-Math-1.5B and Qwen2.5-Math-7B respectively. 
Under GRPO settings, the improvements were $13.80\%$ and $2.44\%$ for the same models. 
Notably, our CCL training strategy yielded consistent performance gains across all test subsets, demonstrating its significant enhancement of model generalization capabilities.
Appendix \ref{appendix:result} also presents the overall performance changes of the CCL method on the test set during the multi-stage training process.

\subsection{Ablation Study}
To evaluate the effectiveness of our method's components, we perform ablation studies on three key aspects: sample difficulty definition, difficult sample processing, and data mixing strategies. 
These studies isolate each design choice's contribution to model performance and training stability.

\subsubsection{Sample Difficulty Definition} 
Previous researches have relied on predefined difficulty labels from the MATH dataset to construct curriculum learning data. 
However, we propose that a more effective approach is to customize difficulty labels based on each model's actual performance on samples before constructing curriculum datasets. We conducted comparative experiments using both approaches to partition training data. 

\begin{figure}[t]
  \centering
  \includegraphics[width=\linewidth]{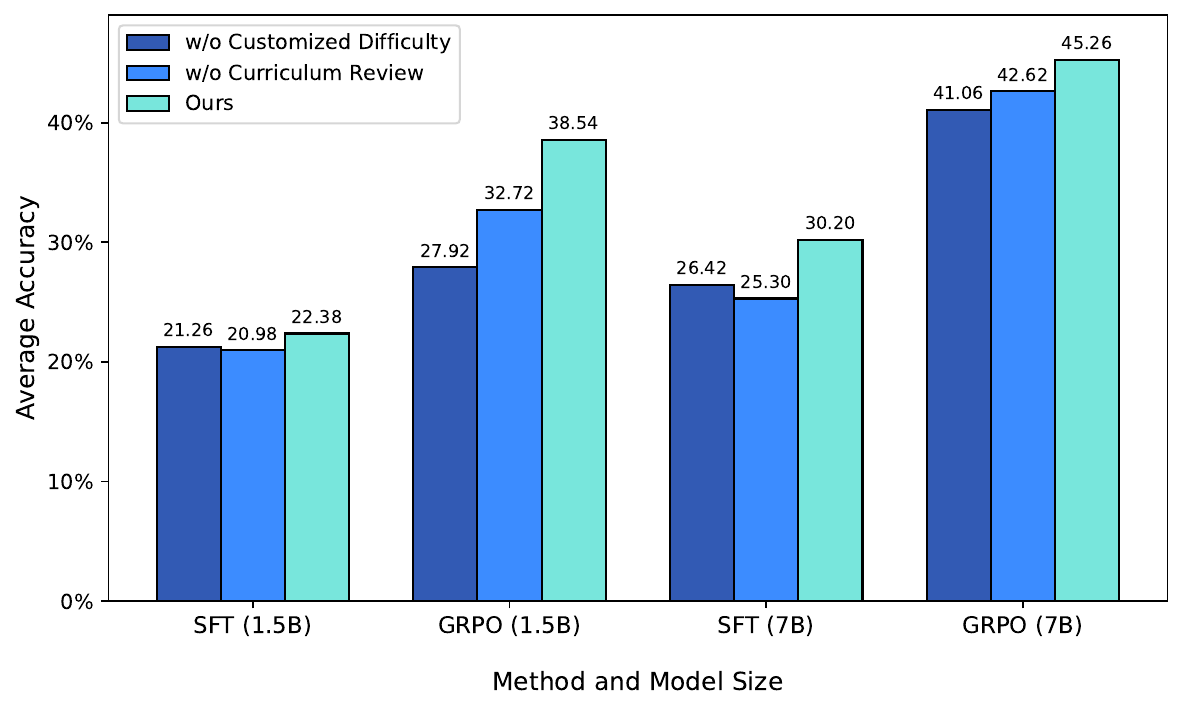} 
\caption{Comparison of model performance after training with data partitioned using different sample difficulty definition methods and across various data mixing strategies.}
  \label{fig:ablation}
  \vspace{-4mm}
\end{figure}

As shown in Figure \ref{fig:ablation}, our customized difficulty definition yields superior performance across models of varying sizes and different post-training methods, demonstrating consistent advantages over the predefined difficulty categorization. 
This empirically validates that difficulty labels tailored to specific model capabilities lead to more effective curriculum learning than predefined difficulty metrics.

\subsubsection{Difficult Sample Processing}
At this part, we conduct an ablation study to test whether continuous training on excessively difficult samples degrades model performance.
We compare three difficult sample processing methods using GRPO: (1) retaining difficult samples, (2) discarding difficult samples, and (3) applying our "Guided Prompting" approach to adapt difficult samples with strategic hints.
Results in Table \ref{tab:challenging sample handling} confirm that learning from overly challenging samples indeed harms model performance. 
Discarding difficult samples outperforms retaining them, validating this hypothesis. 
This effect is more severe for weaker models like Qwen2.5-Math-1.5B, which encounter more unsolvable samples and suffer greater performance degradation.

Rather than waste valuable data through direct removal, our "Guided Prompting" method repurposes difficult samples by providing hints that bring problems within the model's solvable range. Experimental results demonstrate this approach successfully recovers difficult samples while substantially improving model performance, establishing it as the optimal strategy for enhancing mathematical reasoning capabilities.

\setlength{\tabcolsep}{6pt} % 设置表列间距

\begin{table}[htbp]
\centering
\small
\begin{tabular}{llc}
\hline
\textbf{Model Size} & \textbf{Processing Strategy} & \textbf{Avg.} \\ 
\hline

\multirow{3}{*}{\makecell[l]{\textbf{1.5B}}}
 & Retaining Difficult Samples & $26.36$ \\
 & Discarding Difficult Samples & $37.46$ \\
 & Adapting Difficult Samples & $\bm{38.54}$ \\
\hline

\multirow{3}{*}{\makecell[l]{\textbf{7B}}}
 & Retaining Difficult Samples & $41.34$ \\
 & Discarding Difficult Samples & $44.86$ \\
 & Adapting Difficult Samples & $\bm{45.26}$ \\
\hline

\end{tabular}
\caption{Comparison of model performance across three difficult sample processing methods}
\label{tab:challenging sample handling}
\end{table}

\subsubsection{Data Mixing Strategy}
Just as students need to periodically review previously mastered knowledge during their learning process, we believe that models undergoing staged curriculum learning require similar reinforcement of previously acquired content. 
Therefore, we design two distinct data mixing strategies for comparative analysis.
The first approach, termed "Naive Curriculum," provides models with samples corresponding only to the current difficulty level at each training stage. 
The second approach, called "Curriculum Review," incorporates a small proportion of easier samples during later training stages, allowing the model to revisit previously learned material.

Experimental results in Figure \ref{fig:ablation} demonstrate that Curriculum Review data mixing strategy achieves superior performance, confirming that providing models with previously learned content during later training stages is crucial for preventing catastrophic forgetting. 
This finding underscores the importance of maintaining access to foundational knowledge throughout the curriculum learning process.

\section{Conclusions}
In conclusion, we presented Customized Curriculum Learning (CCL), a novel post-training framework that systematically constructs model-adaptive curriculum sequences and transforms difficult samples through guided prompting to enhance large language models' mathematical reasoning capabilities. 
Our comprehensive experiments demonstrated that models trained with CCL significantly outperform those using uniform training approaches across multiple mathematical reasoning benchmarks, with consistent improvements observed in both supervised fine-tuning and reinforcement learning paradigms. 
By effectively integrating curriculum learning into large language model training through model-specific difficulty customization and guided prompting, our work substantially improves sample utilization and model performance, advancing more effective training methodologies for large-scale language models.

\section*{Limitations}
Despite the promising results of our work, several limitations warrant acknowledgment.
While our study focuses on mathematical reasoning,  we see great potential in extending the CCL framework to other domains such as logical reasoning, code generation, and natural language inference, allowing us to further investigate its generalizability across diverse task types.
Furthermore, although our current study applies CCL within specific post-training paradigms, such as supervised fine-tuning and GRPO, we recognize that combining CCL with other post-training strategies—like PPO and broader reinforcement learning techniques—remains an open direction. Exploring these combinations may further reveal the full potential of the CCL framework in enhancing model learning dynamics.

% \section*{Acknowledgments}
\bibliography{custom}

\clearpage
\appendix

\section{Training Details}
In this chapter, we will provide a detailed description of the construction of curriculum datasets in the multi-stage training process, while also presenting a comprehensive overview of the hyperparameters utilized during both training and testing procedures.
\subsection{Constructing Dataset}
\label{appendix:data}
In the process of constructing the curriculum dataset, we need to feed all training set samples into the pre-trained model for inference and evaluate the model's accuracy on each sample. 
To ensure that the evaluation results are as reliable as possible while not causing excessive computational overhead, for each question in the dataset, we use the VLLM framework to generate $16$ responses from the model, extract predictions from these responses using appropriate scripts, and compare them with golden answers to determine the correctness of the generations. To fully harness the model's potential, we did not adopt a greedy decoding strategy to generate responses, but instead set the temperature to $0.7$, generating responses through sampling.

After calculating the model's accuracy on the samples through the above steps, we sort the samples and divide them into $3$ equal parts according to quantity. 
The top $1/3$ with the highest accuracy are classified as simple samples, used for the first stage of model training. 
The bottom $1/3$ with the lowest accuracy are classified as difficult samples, used for the final stage of model training.

\begin{figure*}[!t]
  \centering
  \includegraphics[width=\linewidth]{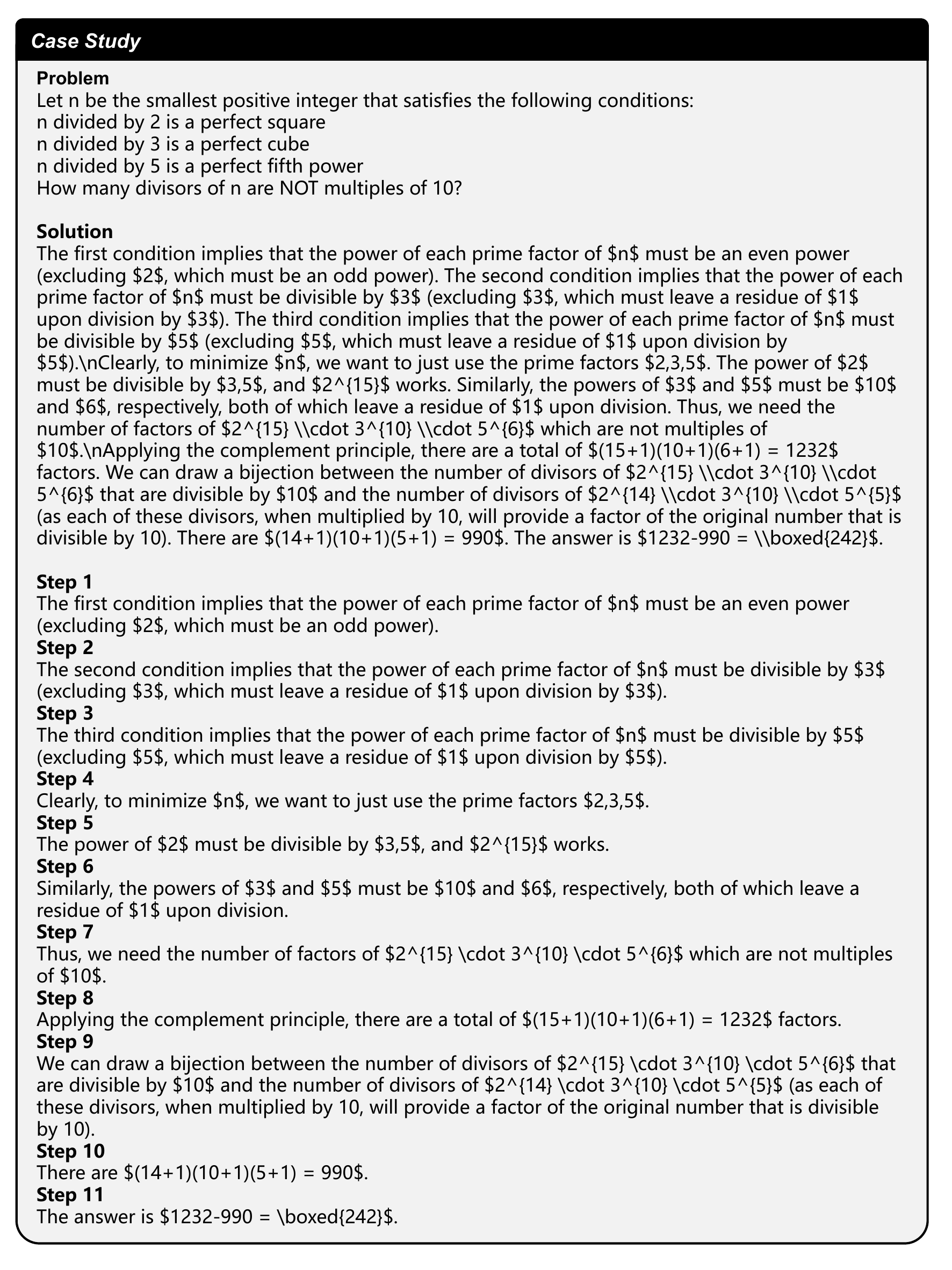}
  \caption{Decomposition of reference answers into step-by-step solution.}
\label{fig:demo}
\end{figure*}

In addition, for particularly challenging samples, we employed a "Guided Prompting" approach to reduce the difficulty for the model. 
Specifically, we first collected reference answers for these difficult samples, then segmented these reference answers into step-by-step reasoning processes, as illustrated in Figure \ref{fig:demo}.
Finally, we selected a small portion of the prefix combined with the original question as input to assist the model in solving problems more effectively.

\subsection{Training HyperParameters}
\label{appendix:hp}
\noindent\textbf{SFT.} 
We conducted our experiments using bf16 precision under the DeepSpeed framework with zero-2 configuration. We set per\_device\_train\_batch\_size to $1$ and gradient\_accumulation\_steps to $4$, employing a cosine lr\_scheduler with warmup set to $0.1$ and max length set to $2048$. For the Qwen2.5-Math-1.5B model, we used a learning rate of $5e-6$ and trained for $3$ epochs. For the Qwen2.5-Math-7B model, we used a learning rate of $1e-5$ and trained for $3$ epochs.

\noindent\textbf{GRPO.} 
We conducted our experiments using bf16 precision under the DeepSpeed framework with zero-2 configuration. 
We set per\_device\_train\_batch\_size to $16$ and gradient\_accumulation\_steps to $8$, employing a cosine lr\_scheduler with warmup set to $0.1$ and beta to $0.04$, num\_generations to $7$, max\_prompt\_lengtht to $512$ and max\_completion\_length $1024$. 
For the Qwen2.5-Math-1.5B model, we used a learning rate of $3e-6$ and trained for $6$ epochs. For the Qwen2.5-Math-7B model, we used a learning rate of $3e-6$ and trained for $4$ epochs.

\section{Evaluation Details}
\label{appendix:eval}
During the testing process, to ensure the stability of test results, all methods employed a greedy decoding strategy with top\_p set to $0.95$, and used "</answer>" as a stop word to truncate the generated content.

\section{Prompt Details}
\label{appendix:prompt}
\begin{figure*}[!t]
  \centering
  \includegraphics[width=\linewidth]{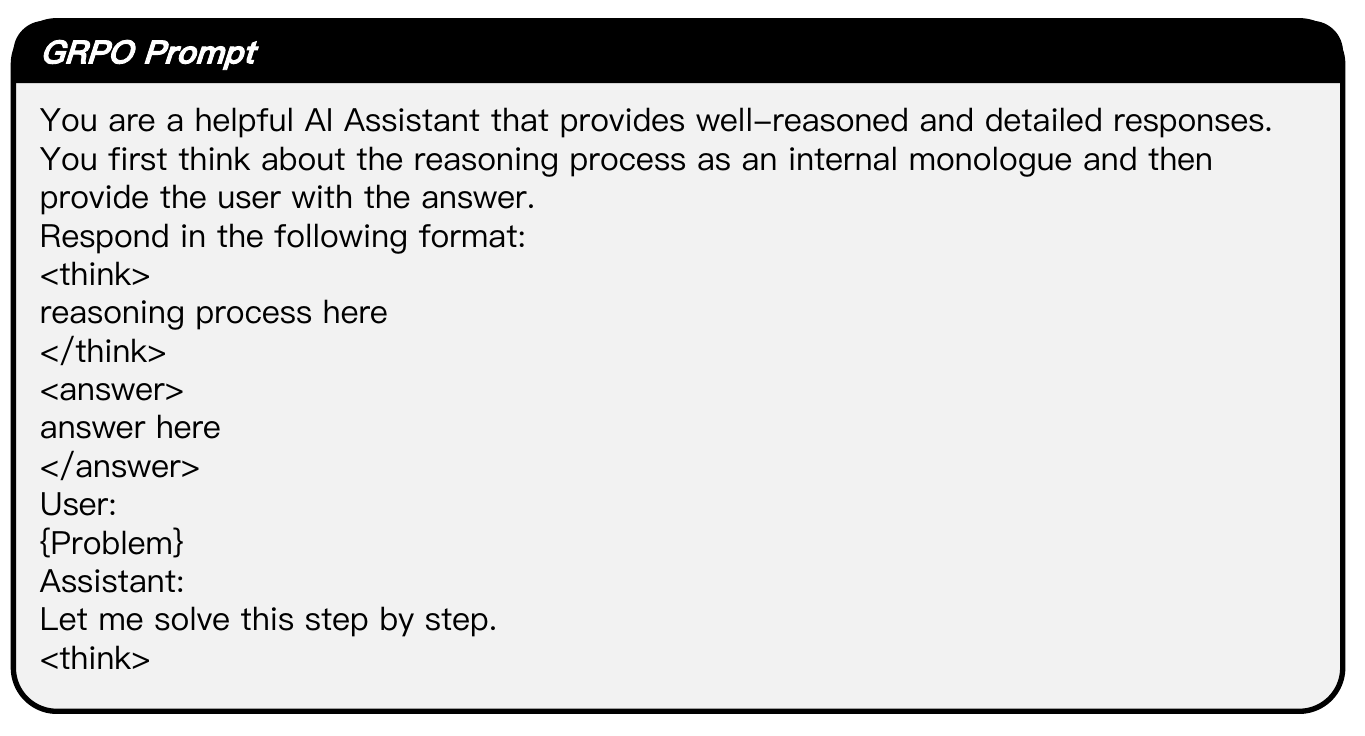}
  \caption{Prompt Used in GRPO.}
\label{fig:grpo_prompt}
\end{figure*}

\begin{figure*}[!t]
  \centering
  \includegraphics[width=\linewidth]{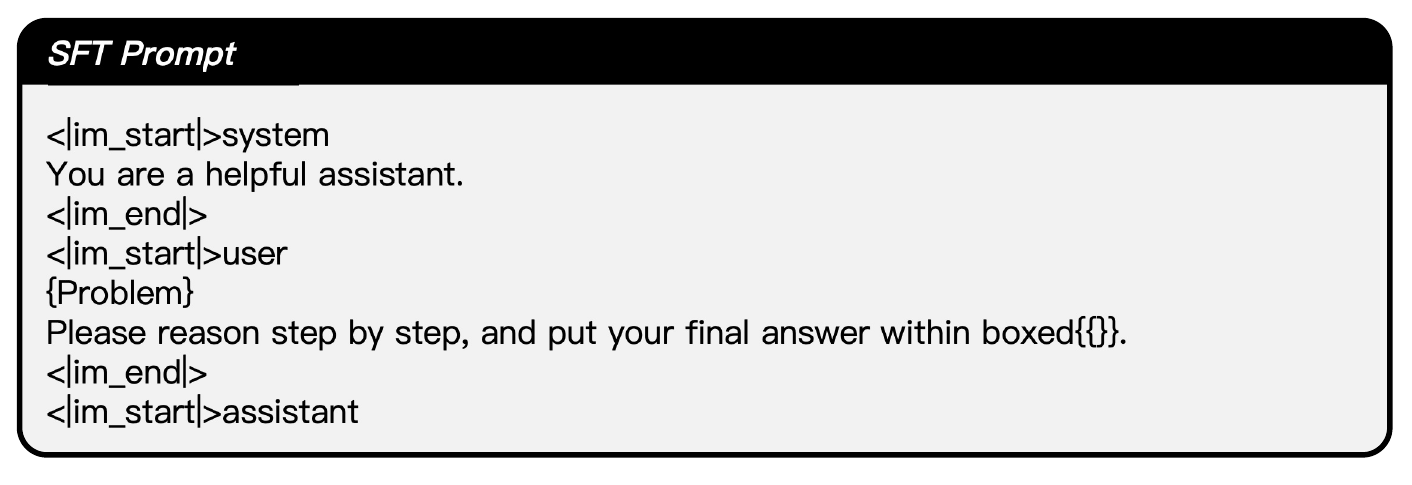}
  \caption{Prompt Used in SFT.}
\label{fig:sft_prompt}
\end{figure*}

During both training and testing processes, the data was processed into a conversational format. 
Figure \ref{fig:sft_prompt} and Figure \ref{fig:grpo_prompt} demonstrate the prompts we used during the SFT and GRPO processes respectively. 
After training the models using their respective methods, we employed the corresponding prompts during testing as well. 
Additionally, during the GRPO training process, besides adding the User's description, we also appended part of the Assistant's content prefixed with the special token "<think>". 
This approach helps the model quickly learn format compliance during the reinforcement learning process, greatly enhancing the stability of the model's reinforcement learning.

\section{Result Details}
\label{appendix:result}
\begin{figure*}[!t]
  \centering
  \includegraphics[width=\linewidth]{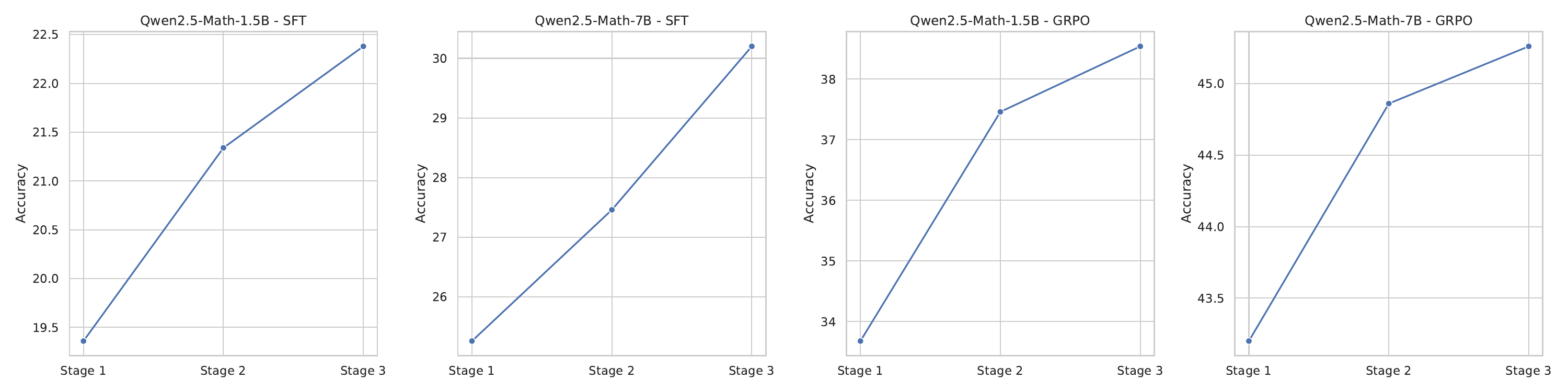}
  \caption{Performance Across Training Stages Using CCL.}
\label{fig:stage_acc}
\end{figure*}

In this section, we demonstrate the overall performance changes on the test set when applying CCL to Qwen2.5-Math-1.5B and Qwen2.5-Math-7B using supervised fine-tuning and reinforcement learning methods for multi-stage training. 
As shown in Figure \ref{fig:stage_acc}, our CCL method continuously improves in performance as training iterations progress.

\label{sec:appendix}

\end{document}